\documentclass[journal]{article}
\usepackage[utf8]{inputenc}
\usepackage{amsmath, amssymb, amsfonts}
\usepackage{graphicx}
\usepackage{booktabs}
\usepackage{hyperref}

\title{Looking through Glass Box}
\author{Alexis Kafantaris \\ \textit{Athens University of Economics and Business}}
\date{February 2026}

\begin{document}

\maketitle

\begin{abstract}
This paper is about a neural implementation of the fuzzy cognitive map, the FHM, and corresponding evaluations. Firstly, a neural net has been designed to behave the same way that an FCM does; as inputs it accepts many fuzzy cognitive maps and propagates them in order to learn causality patterns. Moreover, the network uses langevin differential Dynamics, which avoid overfit, to inverse solve the output node values according to some policy. Nevertheless, having obtained an inverse solution provides the user a modification criterion. Having the modification criterion suggests that information is now according to discretion as a different service or product is a better fit. Lastly, evaluation has been done on several data sets in order to examine the networks performance.
\end{abstract}

\section{Introduction}
In this paper a glass box neural net architecture is presented and analyzed. The glass box architecture is a design that provides learning constraints to satisfy physics or causal directions according to requirements. Originally, the system was meant as a black box, however, fathoming it produced a result that resemble a glass box more.

Moreover, this architecture is comprised of a simple, yet elegant, mathematical model that uses fusion as its core mechanism. On the other hand, this model achieves understanding of the problem by imposing constrains over its iterations and attempting to replicate known knowledge on its results. That is achieved by emphasizing core logical aspects of learning through the model equations.

In addition, the system is evaluated for various datasets and situations in regards practical use. On that note, there are several observations as well as an example of the MPG data set and fuzzy logic inferences that can be drawn. Lastly, some comments are made about this particular architecture as well as some concluding remarks. This article is organized slightly differently than a conventional paper; at first there is literature review, and some comments about methodology, and then after that there is a foundation for the mathematical model, as well as the discussion and the results. Finally there are some concluding remarks as well as suggestions about the feature.

\section{Literature review}
To begin, this paper is composed in order to address the core issues of fuzzy optimization of information transmission of service design processes; nonetheless, the information that is included here acts as a collateral benefit. Having stated that, on an attempt to model the previous problem information transmission has to be examined. To examine information transmission one has to model the service as a whole, i.e. services as a network; and that is achieved using a multiplex as in many other problems.

At a first glance, depending on the problem at hand research ventures use state of the art technology [12] en par with classical optimization; while some seem to lean toward genetic algorithms there are still some that put their faith in the FCM [8] [7]. One interesting note is the multiplex [14] which models arrays of information. Another interesting aspect of the multiplex is the alternative capabilities that it provides in respect to hierarchy. There is an implicit connection between optimizing the static structure and a multiplex. For example the ant colony optimization [2] [3] et al. involves an interesting perspective that addresses the dynamics of a structure using the pheromone as well the ant colony. More importantly the pheromone is used in combination with the multiplex colony and it creates powerful evolutionary system [5] [9].

Moreover, graph theory addresses both multiplexes as well as statics of the structure [5] [6] [1] in an interesting way. The nested tree structure for example is aforementioned to successfully encapsulate the FCM in a solid way. For that, a research gap has been identified regarding alternatives to imitate the FCM. And although there are numerous attempts to bridge the logic or the data of information systems, there is still a pending question about implication modeling and neural nets [4]. The core of the research is that one can potentially bridge a gap in service design processes using soft computing [15] or more specifically neural nets [13]. On the one hand there are fuzzy cognitive maps and other variant some mathematical tools [8] to help determine the same problems. More specifically, there is not a multiplex that through the use of soft computing imitates the FCM, while this can be done using a neural net. Attempts have been made for inverse FCM solution, one of the relevant problems, using a multitude of methods like q-FCM [7] et al. and so one is attempted using the FHM.

\section{Methodology}
The methodology is mainly comprised of a mathematical model as well as its implementation. The model maps directly to python code and vice versa; it is noted that this way it is easier to expound on a heavy and theoretical idea while maintaining the integrity of the object. Furthermore, the core of the solution is a neural net that works like a glass box; that is, a mathematical object that learns the structure of information and while minimizing the objective loss it also understands information that is fed to it. It is very much alike a physics informed neural net in a sense [10].

\subsection{Model Definition}
\textbf{Inputs}: $A, X$, Where: $X$ is the input data, and $A$ is the adjacency matrix.

\textbf{Encoder}:
\begin{equation}
H_{0}=tanh(X_{0}W_{1}+b_{1})W_{2}+b_{2}
\end{equation}
where: this is the inner most layer and it is used to calculate-guess the node influences-weights according to the graph.

\textbf{Metrics Projection}:
\begin{equation}
Y_{m}=Softsign(ReLU(H_{0}W_{m1}+b_{m1})W_{m2}+b_{m2})
\end{equation}
where: this is the outer layer that loss function is implemented; it encapsulate the inner layer as a latent dimension and is the loss target.

\textbf{Parameters}:
$W_{m1}\in\mathbb{R}^{D_{latent}\times4}$, $b_{m1}\in\mathbb{R}^{4}$, $W_{m2}\in\mathbb{R}^{4\times1}$, $b_{m2}\in\mathbb{R}^{1}$

\textbf{Mini-FCM Objective}:
\begin{equation}
min\sum w-G
\end{equation}
where: the $w$ is the $i,j$ weight from each node and the $G$ is the latent embeddings. The mini Fcm is used as a normalizer fusion function.

\subsection{Process}
Stacked Node Embeddings and:
\begin{equation}
H_{curr}=Encoder(X)
\end{equation}
\begin{equation}
H_{prop}=MiniFCM(H_{curr})=MiniFCM(Encoder(X))
\end{equation}
\begin{equation}
H_{t+1}=tanh(H_{t}+H_{prop})+tanh(5\cdot H_{cur})
\end{equation}
\begin{equation}
H_{t}^{0}=H_{curr}+sign(H_{curr})
\end{equation}
That enforces adjacency constrain on each Fused matrix turning the black box into a a transparent one. That operator is both during the learning of the next operation and also during the selection. The difference is that tanh is differentiable so during propagation it is possible to learn the matrix while at the end it is enforced.

Helping definitions At each iteration $t$, the model is selected based on a criterion and past embeddings $S$, one that relies on ground truth data, whether it is steady or not.

\textbf{Node Force}: $E_{t}=||H_{curr}^{(t)}||_{2}$

\textbf{Transitive Alignment}: $\Gamma_{t}=E_{t}\cdot(A\odot|S|)^{\top}$

\textbf{Causal Score}: $\mathcal{S}_{t}=mean(E_{t}+\Gamma_{t})$

\textbf{Propagation Operator}:
$\Phi(H_{t})=tanh(H_{t}+H_{prop})+tanh(5\cdot H_{curr})$ \\
$H_{t+1}=\Phi(H_{t})$ \\
This operator is used to update $H$ every time a new step is required for the model.

\textbf{Final Iterated State}:
\begin{equation}
H_{T}=\Phi\circ\Phi\circ\cdot\cdot\cdot\circ\Phi(H_{0})
\end{equation}
The $H_t$ is calculated by the previous formula as aforementioned.

\textbf{Output Logic Gate}:
\begin{equation}
H_{final}=H_{T}+sign(H_{T})
\end{equation}
This acts as an adjacency constrain as $H_t$ is strictly belonging in [0,1].

\textbf{Best-State Selection}:
\begin{equation}
\hat{H}_{t+1}=\{\begin{matrix}H_{t+1},&if~S_{t}>S_{perf}\\ \hat{H}_{perf},&otherwise\end{matrix}
\end{equation}
Here is the criterion of termination, it is noted that steps is another criterion.

\textbf{Total Loss}:
\begin{equation}
\mathcal{L}_{tune}=||Y_{m}^{(k)}-v||^{2}
\end{equation}
\begin{equation}
v=\frac{1}{N}\sum_{i=1}^{N}(\frac{1}{d_{m}}\sum_{j=1}^{d_{m}}DATA_{i,j}^{(m)})
\end{equation}
\begin{equation}
\mathcal{Y}_{m}^{(k)}= Softsign (ReLU((H_T + sign(H_T))W_{m1}+b_{m1})W_{m2}+b_{m2})
\end{equation}

Therefore:
\begin{multline}
\mathcal{L}_{tune}=||Softsign(ReLU((H_{T}+sign(H_{T}))W_{m1}+b_{m1})W_{m2}+b_{m2})_{k} \\
-\frac{1}{N}\sum_{i=1}^{N}(\frac{1}{d_{m}}\sum_{j=1}^{d_{m}}X_{i,j}^{(m)})||^{2}
\end{multline}

Now for the latent space optimization:
\begin{equation}
\mathcal{L}_{tune}=||Softsign(ReLU([...ENCODER+FCM...])W_{m1}+b_{m1})W_{m2}+b_{m2})_{k}-v||^{2}
\end{equation}
\begin{equation}
\mathcal{L}_{tune}=\|Softsign(ReLU(\|
\end{equation}
\begin{equation}
[tanh(\frac{H_{T-1}}{Encoder Accumulation}+\frac{H_{prop}^{(T-1)}}{Mini-FCM})+...]W_{m1}+b_{m1})W_{m2}+b_{m2})_{k}-v||^{2}
\end{equation}
\begin{multline}
\mathcal{L}_{tune}=||Softsign(ReLU(([tanh(H_{T-1}+H_{prop}^{(T-1)})+tanh(5\cdot H_{T-1})] \\
+sign(tanh(H_{T-1}+H_{prop}^{(T-1)})+tanh(5\cdot H_{T-1})))W_{m1}+b_{m1})W_{m2}+b_{m2})_{k} \\
-\frac{1}{N}\sum_{i=1}^{N}(\frac{1}{d_{m}}\sum_{j=1}^{d_{m}}X_{i,j}^{(m)})||^{2}
\end{multline}
Which allows the $H_t$ to be optimal. In other words, the system minimizes the loss while it tries to optimize the equation for H matrix. Metrics heads act as a information projection, then are minimized according to targets. By minimizing outer layer according to targets the encoder+fcm are forced to learn the weights and causality. Lastly, this way the system achieves minimization only when the H matrix has information of to sign and adjacency output.

\section{Inverse solution}
\textbf{Inverse Problem Initialization \& Flow}: \\
FCM Matrix \& Masks \\
$W=FCM(H_{stored})$ \\
$M_{valid}=A_{mask},$ $M_{forbidden}=1-M_{valid}$ \\
The network assumes mask for valid paths and for invalid paths; these masks are later used to evaluate loss and topology.

\textbf{Parameter Initialization}: $\theta_{0}=0$. Optimized via Stochastic Gradient Descent (SGD) with momentum.

\textbf{Masks}: For the raw input parameters $x\in[0,1]$, $x=\sigma(x)$, the FCM weights are used in accordance with masks to calculate loss:
\begin{equation}
F_{valid}=(W\odot M_{valid})\sigma(x)
\end{equation}
\begin{equation}
F_{forbidden}=(W\odot M_{forbidden})\sigma(x)
\end{equation}

\textbf{Stochastic Activation (Simulated Annealing) Inversion}: In stept of the total steps $T$, noise $\sim \mathcal{N}(0,0.01^{2})$ is applied to escape local minima:
\begin{equation}
Y_t = \begin{cases} \sigma(F_{valid} + F_{forbidden} + \epsilon), & \text{if } t < \frac{T}{2} \\ (F_{valid} + F_{forbidden}), & \text{otherwise} \end{cases}
\end{equation}

\textbf{Objective Function}: For each node the $y$ calculated using a mask as soft constrain, and then a topological constrain as well as target data loss constrain are imposed.
\begin{equation}
\mathcal{L}_{target}=100\cdot(y_{t,i}-v)^{2}
\end{equation}
\begin{equation}
\lambda_{t}=\lambda_{soft}(1+\frac{t}{T})
\end{equation}
\begin{equation}
\mathcal{L}_{topology}=\lambda_{t}||F_{forbidden}||_{2}
\end{equation}
\begin{equation}
\mathcal{L}_{total}=\mathcal{L}_{target}+\mathcal{L}_{topology}
\end{equation}

\textbf{Output}: Finally the inverse comes from minimizing total loss $\mathcal{L}_{total}$ to $y$ target from using the following activation:
$Output = \sigma(Wo( ))$ i.e. minimize what the model calculates to what the inputs calculate.
Minimize $:Y-\sigma(W\sigma(x))$

\section{Experimental Results and Discussion}
The neural network has been examined on roughly ten data sets; firstly synthetic data set for a smart City has been evaluated at 9, 14, 19, and 24 node sizes. These all have randomly generated data, and nonetheless the models performance is exceptional or even stable. Additionally, the sachs [11] problem which is a protein has also been evaluated for 11 and 25 nodes. It is verified that it works here and is stable. Additionally, the IEEE bus topology as well as the auto MPG data set have been included to solidify the claims of a functioning prototype.

\begin{table}[h]
\centering
\caption{Summary of Causal Evaluation Experiments}
\begin{tabular}{lccc}
\toprule
Experiment / Topology & Nodes & Direct Edge Acc. & Transitive Chain Acc. \\
\midrule
Base Urban Policy & 9 & $99.29\%\pm2.20\%$ & $99.38\%\pm1.92\%$ \\
Extended Urban Policy & 14 & $90.89\%\pm3.75\%$ & $85.91\%\pm6.06\%$ \\
Ministry Urban Policy* & 19 & $87.50\%\pm5.49\%$ & $74.64\%\pm6.05\%$ \\
Expanded Urban Policy & 24 & $86.49\%\pm3.69\%$ & $76.13\%\pm4.87\%$ \\
Sachs Protein Network & 11 & $76.83\%\pm8.20\%$ & $75.00\%\pm13.46\%$ \\
Sachs Protein Network & 25 & $74.60\%\pm8.56\%$ & $63.50\%\pm9.10\%$ \\
Auto MPG (Mech.) & 6 & $79.29\%\pm7.11\%$ & $83.12\%\pm7.93\%$ \\
IEEE Power Grid & 14 & $73.88\%\pm11.48\%$ & $81.88\%\pm20.87\%$ \\
\bottomrule
\end{tabular}
\end{table}

*Note: Although the section header states "20 node", the code defines 19 distinct nodes.

One of the interesting datasets was the MPG due to the fact that the system examined real data. A major challenge was to create a working system and this way it was addressed. Furthermore, the system worked great; irrespective of real data or synthetics it provided good and accurate predictions on the cross validations that were done. More problems can now be traced and solved especially for the real data. The math concept is interesting and it is aligned with the code. Some other more sophisticated variants were tested, such as one that takes into account parent-child relationship at depth one and uses that as a criterion, but with no additional benefit whatsoever. Another interesting idea was to use full dynamics for node child-parent relationship instead of recursive depth, yet the results were roughly the same if not worse.

Here the first thing that is observed is that according to input specifications a rental company is able to provide a different cars according to discretion. Someone walks in and asks for a good car, and good means low cost and high quality for which there are node entities to search; now instead of relying solely on cosine metrics one uses fuzzy logic to define the membership of good and then search for that similarity through the inverse solution. The other thing is that there are good and bad evaluation folds; one only cares about the good evaluation fold for the best evaluation folds as that fold understands the solution better. Additionally, the system knows the general pattern of the solution from which he can tell whether or not the inverse policy is right or wrong. Selecting the best inversion results in the most accurate policy. Yet that is a different topic; for now the issue is mostly about imitating a fuzzy cognitive map using a neural net.

\section{Conclusion}
To conclude, this paper describes the implementation of a glass-box architecture as well as some examples. A research gap in neuro-symbolic fuzzy computing is identified and then a multiplex design is used to address it. Through proposed architecture, which is transparent, one can emulate fuzzy cognitive maps and assert some interesting properties like differentiability. Lastly, interest piques as the proposed architecture is transparent; while minimizing the loss one forces the model to behave according to some wanted properties. And that is seemingly a good shift in logic. During the design process of this model it became evident that such an idea works great; for this reason it is hoped that future neural net trends shift towards the glass box.
\pagebreak
\section*{Acknowledgments}
It is acknowledged that this paper is part of a PhD dissertation, fuzzy optimization of information transmission in service design process currently done in (Athens University of Economics and Business) AUEB. It is also written in correspondence with Dr. Dimitris Kardaras which was the supervisor of the specific subject in AUEB and also interested in service optimization.

\end{document}